\documentclass[sigconf]{acmart}
\usepackage{booktabs} 
\usepackage{amssymb}
\usepackage{hyphenat}
\usepackage{xspace}
\usepackage{subfigure}
\usepackage{url}
\usepackage{amsmath}
\usepackage{multirow}
\usepackage{minibox}
\usepackage{braket}
\usepackage[show]{chato-notes} 

\newcommand{\hide}[1]{}

\newcommand{\xhdr}[1]{{\bf #1.}}

\newcommand{\ie}{{i.e.}\xspace}
\newcommand{\eg}{{e.g.}\xspace}
\newcommand{\cf}{{cf.}\xspace}

\newcommand{\vs}{{vs.}\xspace}

\newcommand{\name}{Cr5\xspace}
\newcommand{\cl}{crosslingual\xspace}
\newcommand{\Cl}{Crosslingual\xspace}
\newcommand{\baseline}{\textsc{AdvRefine}\xspace}

\newcommand{\cpt}[1]{\textsc{\MakeLowercase{#1}}}
 
\newcommand{\Secref}[1]{Sec.~\ref{#1}}

\newcommand{\Eqnref}[1]{Eq.~\ref{#1}}

\newcommand{\Tabref}[1]{Table~\ref{#1}}

\newcommand{\Figref}[1]{Fig.~\ref{#1}}

\newcommand{\affilSize}{9pt}
\newcommand{\authorbox}[3]{
\minibox[c]{
#1\\
{\fontsize{\affilSize}{\affilSize}\selectfont{}#2}\\
{\fontsize{\affilSize}{\affilSize}\selectfont{}#3}
}
}

\providecommand{\R}{\mathbb{R}} 

  \providecommand{\bb}{\mathbf{b}}

  \providecommand{\hh}{\mathbf{h}}

  \providecommand{\mm}{\mathbf{m}}

  \providecommand{\uu}{\mathbf{u}}
  \providecommand{\vv}{\mathbf{v}}
  \providecommand{\ww}{\mathbf{w}}
  \providecommand{\xx}{\mathbf{x}}
  
  \providecommand{\zz}{\mathbf{z}}
  \providecommand{\vmu}{\mathbf{\mu}}
  \providecommand{\vzeta}{\mathbf{\zeta}}
  
  \providecommand{\mB}{\mathbf{B}}

  \providecommand{\mH}{\mathbf{H}}
  \providecommand{\mI}{\mathbf{I}}

  \providecommand{\mL}{\mathbf{L}}
  \providecommand{\mM}{\mathbf{M}}

  \providecommand{\mP}{\mathbf{P}}
  \providecommand{\mQ}{\mathbf{Q}}

  \providecommand{\mT}{\mathbf{T}}
  
  \providecommand{\mV}{\mathbf{V}}
  \providecommand{\mW}{\mathbf{W}}
  \providecommand{\mX}{\mathbf{X}}
  \providecommand{\mY}{\mathbf{Y}}
  \providecommand{\mZ}{\mathbf{Z}}
  \providecommand{\mPhi}{\mathbf{\Phi}}
  \providecommand{\mLambda}{\mathbf{\Lambda}}
  \providecommand{\mSigma}{\mathbf{\Sigma}}

\begin{document}

\copyrightyear{2019}
\acmYear{2019}
\setcopyright{acmlicensed}
\acmConference[WSDM '19]{The Twelfth ACM International Conference on Web Search and Data Mining}{February 11--15, 2019}{Melbourne, VIC, Australia}
\acmBooktitle{The Twelfth ACM International Conference on Web Search and Data Mining (WSDM '19), February 11--15, 2019, Melbourne, VIC, Australia}
\acmPrice{15.00}
\acmDOI{10.1145/3289600.3291023}
\acmISBN{978-1-4503-5940-5/19/02}

\settopmatter{printacmref=false, printfolios=false}


\title[Crosslingual Document Embedding as Reduced-Rank Ridge Regression]{Crosslingual Document Embedding\\as Reduced-Rank Ridge Regression}

\author{
\authorbox{Martin Josifoski}{EPFL}{martin.josifoski@epf\/l.ch}
\authorbox{Ivan S. Paskov}{MIT}{ipaskov@mit.edu}
\authorbox{Hristo S. Paskov}{BlackRock}{hristo.paskov@blackrock.com}
\authorbox{Martin Jaggi}{EPFL}{martin.jaggi@epf\/l.ch}
\authorbox{Robert West}{EPFL}{robert.west@epf\/l.ch}
}

\renewcommand{\shortauthors}{M. Josifoski, I. S. Paskov, H. S. Paskov, M. Jaggi, R. West}

\begin{abstract}
There has recently been much interest in extending vector-based word representations to multiple languages, such that words can be compared across languages. In this paper, we shift the focus from words to documents and introduce a method for embedding documents written in any language into a single, language\hyp independent vector space. For training, our approach leverages a multilingual corpus where the same concept is covered in multiple languages (but not necessarily via exact translations), such as Wikipedia. Our method, \textit{\name} (Crosslingual reduced\hyp rank ridge regression), starts by training a ridge\hyp regression--based classifier that uses language\hyp specific bag-of-word features in order to predict the concept that a given document is about. We show that, when constraining the learned weight matrix to be of low rank, it can be factored to obtain the desired mappings from language\hyp specific bags-of-words to language\hyp independent embeddings. As opposed to most prior methods, which use pretrained monolingual word vectors, postprocess them to make them crosslingual, and finally average word vectors to obtain document vectors, \name is trained end-to-end and is thus natively crosslingual as well as document\hyp level. Moreover, since our algorithm uses the singular value decomposition as its core operation, it is highly scalable. Experiments show that our method achieves state-of-the-art performance on a crosslingual document retrieval task. Finally, although not trained for embedding sentences and words, it also achieves competitive performance on crosslingual sentence and word retrieval tasks.
\end{abstract}

\maketitle

{\fontsize{8pt}{8pt} \selectfont
\textbf{ACM Reference Format:}\\
Martin Josifoski, Ivan S. Paskov, Hristo S. Paskov, Martin Jaggi, and Robert West.
2019.
Crosslingual Document Embedding as Reduced-Rank Ridge Regression.
In
\textit{The Twelfth ACM International Conference on Web Search and
Data Mining (WSDM '19), February 11--15, 2019, Melbourne, VIC, Australia.}
ACM, New York, NY, USA, 9 pages. \url{https://doi.org/10.1145/3289600.3291023}
}

\section{Introduction}
\label{sec:intro}

This paper addresses the problem of representing documents written in any language in a language\hyp invariant manner such that documents become seamlessly comparable across languages without the need for full-fledged machine translation (MT).
Solutions to this problem would be tremendously useful;
\eg, the Web is inherently multilingual, and it is becoming ever more so as Internet usage becomes more widespread across the globe.
Classic search engines, however, even when available in multiple languages, usually only return documents written in the same language as the query,
thus discarding many potentially valuable search results written in other languages.
A language\hyp invariant document representation, on the contrary, would allow us to retrieve resources in any language for queries in any other language.
Beyond information retrieval, further useful applications include \cl transfer learning \cite{buys2016cross}, plagiarism detection \cite{potthast2011cross}, and text alignment \cite{gottschalk2017multiwiki}.

Given recent advances in MT, one way forward would be to translate all documents into a pivot language as a preprocessing step and retrieve from that canonical representation \cite{levow2005dictionary, martino2017cross, hull1996querying}.
This approach is impractical, though, when operating at Web scale, as MT is costly and can have difficulties with resource\hyp poor languages.
One should thus avoid full MT and strive for more lightweight \cl representations.
(MT may still be applied once relevant documents have been retrieved based on such a representation, to the small subset of documents presented to the user.)
    
Various approaches have been proposed for obtaining \cl document representations.
Some represent a document via its relation to concepts in a \cl knowledge base, such as Wikipedia \cite{potthast2008wikipedia} or BabelNet \cite{franco2014knowledge}.
While intuitive and straightforward to implement, these methods are heuristic and not optimized via learning.
Other methods therefore build on recent advances in learned distributed word representations \cite{mikolov2013distributed,pennington2014glove}.
They start from monolingual word embeddings, postprocess them to render them \cl \cite{conneau2017word}, and finally obtain document embeddings by combining the word embeddings of its constituent words into a single vector via operations such as summing or averaging \cite{litschko2018unsupervised,vulic2015monolingual}.
Although this approach has been shown to achieve state of the art on both monolingual and \cl information retrieval tasks \cite{litschko2018unsupervised,vulic2015monolingual},
it is still heuristic in nature, as the process of obtaining \cl word embeddings is decoupled from that of combining word embeddings into document embeddings.
In other words, the embeddings are not optimized explicitly for the document level.

\xhdr{Present work: Crosslingual reduced\hyp rank ridge regression (\name)}
This shortcoming provides the starting point for our work.
We introduce a novel end-to-end training method that directly learns mappings from language\hyp specific document representations to a language\hyp invariant embedding space, without the detour via word embeddings.
The key to our method is to use \cl document alignments, as for instance provided by Wikipedia, which contains articles about the same concept in many languages (but not necessarily via exact translations).
Leveraging such alignments, we formulate a linear classification problem of predicting the language\hyp independent concept that an article is about based on the article's language\hyp specific bag-of-word features.
We show that, when the learned weight matrix is constrained to have low rank, it can be decomposed to obtain the desired mappings from language\hyp dependent bag-of-word spaces to a single, language\hyp invariant embedding space.
Moreover, when using ridge regression as the linear classifier, this problem can be reformulated such that it involves exclusively matrix--vector multiplications, which makes it extremely efficient and scalable to very large vocabulary sizes and corpora of millions of documents---a marked difference from neural\hyp network--based approaches.
We name this model \textit{\name,} for \textit{\textbf{Cr}osslingual \textbf{r}educed\hyp \textbf{r}ank \textbf{r}idge \textbf{r}egression}.

We demonstrate \name's ability to embed texts at various levels of granularity by evaluating it on \cl tasks involving long documents, shorter sentences, all the way to single words.
For crosslingual document retrieval, our method offers very strong improvements over the state of the art;
\eg, given a Danish Wikipedia article, the corresponding Vietnamese article is its closest neighbor in the learned embedding space (containing 200k candidates) in 36\% of cases, while the previously best model \cite{conneau2017word} achieves only 8\%.
While the relative boost is especially high for such resource\hyp poor languages, the absolute performance is highest on resource\hyp rich languages; \eg, the above number becomes 79\% when using Italian\slash English instead of Danish\slash Vietnamese.

We also find that, even for two languages that have no documents in common, their intersections with a third language (e.g., English) allows our method to learn high\hyp quality aligned document representations, which we believe is a first in the literature and has a strong impact on the many low\hyp resource languages for which explicit alignment data is often unavailable.

Finally, we show that \name, while trained on the document level, also gives competitive crosslingual sentence and word embeddings.

Our main contributions are threefold.
First, we rigorously formulate document embedding as a multiclass classification problem (\Secref{sec:rank-reduced classification}).
Second, we show how the problem can be efficiently solved using highly optimized linear algebra techniques (\Secref{sec:Cr5}).
And third, we train an instance of our model on multilingual Wikipedia in 4 languages and demonstrate that it achieves state-of-the-art performance on document retrieval (\Secref{sec:doc retrieval}), and competitive performance on sentence (\Secref{sec:sentence retrieval}) and word (\Secref{sec:word retrieval}) retrieval.


\section{Related work}
\label{sec:relwork}

Recent advances in word embeddings, such as distributed word representations \cite{mikolov2013distributed, pennington2014glove} for a given language, produce features reflecting the semantics of each word, and have had tremendous impact on many downstream applications. While initial methods were monolingual, extensive research has also aimed at transferring semantic word-level similarity across languages \cite{klementiev2012inducing, hermann2014multilingual, mikolov2013exploiting}.

Concurrently, researchers have attempted to bridge the gap between languages in the information retrieval setting \cite{levow2005dictionary, potthast2008wikipedia, cox2008cross}, using various approaches. Translation\hyp based methods work by translating either the query or the target \cite{levow2005dictionary, martino2017cross, hull1996querying}, but this makes them dependent on high\hyp performing machine translation systems, which are heavy-weight and not available for all language pairs. Another line of work considers representing a document via its association strength to a predefined set of concepts from an external knowledge base
\cite{franco2014knowledge,gonzalo1998indexing,potthast2008wikipedia}.
Here, the objective of learning language\hyp invariant document representations requires concept descriptions to be comparable across languages. As Wikipedia articles are linked to language\hyp independent concepts, which are in turn usually described in multiple languages, one of the most prominent members of this class, CL-ESA \cite{potthast2008wikipedia}, is based on Wikipedia.
An issue with CL-ESA and similar methods is the low number of available indexing concepts when embedding multiple languages, especially when including low\hyp resource languages, since the intersection of concepts covered across all languages is substantially smaller.
Although our method exploits Wikipedia's \cl alignment the same way, it only requires a concept to be described in at least two languages in order to use it for training, thus alleviating the above issue.
A third line of research considers documents in their bag-of-word representation and generates document embeddings by combining the constituent word representations by summing or averaging \cite{litschko2018unsupervised, vulic2015monolingual}. Using word embeddings that capture semantics across languages makes systems of this kind inherently \cl. Vuli\'c and Moens \cite{vulic2015monolingual} demonstrate the superiority of this approach over the former two in both monolingual and \cl ad-hoc document retrieval tasks on the CLEF benchmarking dataset.

This latter method, however, relies heavily on good \cl word embeddings.
To obtain them, Mikolov et al. \cite{mikolov2013exploiting} observed similarities in the embedding spaces across languages and proposed a way of exploiting them to learn a mapping from a source to a target space. The \cl signal they consider consists of bilingual word dictionaries. Much follow-up research has focused on exploiting similarities between monolingual spaces, aiming at improving \cl word embeddings while decreasing the level of supervision \cite{ammar2016massively, faruqui2014improving, smith2017offline, artetxe2016learning, artetxe2017learning, lazaridou2015hubness, gouws2015bilbowa, xing2015normalized}.
A recently proposed unsupervised method \cite{conneau2017word} for learning \cl embeddings
achieved state-of-the-art performance in word- and sentence\hyp level evaluations.
Litschko et al. \cite{litschko2018unsupervised} further showed that those embeddings are also superior at document\hyp level \cl information retrieval tasks. Consequently, we will use them as our main baseline.

An alternative approach to merging two monolingual embedding spaces into one bilingual space
is to directly learn embeddings that capture the semantic similarity across languages.
This allows them to interact more freely and facilitates transfer learning between similar languages, leading to better generalization \cite{duong2017multilingual, agic2016multilingual}. As this approach depends on a \cl supervision signal, the main issue becomes the amount of available resources with the required level of supervision.
Contrary to word- or sentence\hyp aligned parallel data, methods that require document\hyp aligned (``comparable'') data have proven promising, by significantly alleviating the problem of scarce resources; \eg, Wikipedia, a large dataset, may be used as a comparable corpus.
S{\o}gaard et al. \cite{sogaard2015inverted} frame word embedding in terms of dimensionality reduction.
They first represent each word via its relation to predefined language\hyp independent concepts and then project the ensuing matrix into a lower\hyp dimensional space via the singular value decomposition.
Since the solution relies solely on a highly optimized linear algebra routine, it scales gracefully.
Vuli\'c and Moens \cite{vulic2015monolingual} exploit document\hyp aligned data through a pseudo\hyp bilingual corpus, obtained by merging the aligned documents in the two languages, shuffling the order of words (thus mixing the languages), and finally applying the standard skip-gram model \cite{mikolov2013distributed} to learn \cl word embeddings.
Here, we propose a novel approach for learning \cl embeddings by framing the problem in a multiclass classification setting, which uses the available data in a highly efficient and scalable manner.


\section{Method}
\label{sec:method}

\subsection{Embedding via reduced-rank classification}
\label{sec:rank-reduced classification}

Every embedding procedure requires information from which to learn the geometry of its embedded points; we focus on the use of class labels, which partition sample points into equivalence classes, to anchor our embedding. Our goal is to find a linear map that preserves this equivalence\hyp class structure by placing points so that ones with the same class label are closer to each other in the embedding than they are to points with different class labels. We begin by describing how a variety of multiclass classification algorithms can be interpreted as providing such linear maps and then show how this general methodology can be applied to our problem of crosslingual embedding. 

To set the stage, suppose we are given data points $\xx_1,\dots,\xx_n \in \R^p$ along with class labels $y_1,\dots,y_n \in \{1,\dots,K\}$ and wish to find separating hyperplanes $\ww_k\in\R^p, b_k \in \R$ for each class $k\in \{1,\dots,K\}$ whereby the ``winner-takes-all'' decision rule
\begin{equation}\label{eq:winner_takes_all}
y(\xx) := \underset{k\in\{1,\dots,K\}}{\text{arg max }} \ww_k^\top \xx  + b_k    
\end{equation}
correctly discriminates points in $\R^p$. A variety of multiclass classification methods including multinomial logistic regression \cite{friedman2001elements}, multiclass support vector machines \cite{weston1998multi, crammer2001algorithmic, lee2004multicategory, friedman2001elements}, and one-\vs-all regularized least squares classification \cite{friedman2001elements} are appropriate in this setting and may be interpreted as finding a $K \times p$ matrix $\mW$ along with a vector of offsets $\bb \in \R^K$ whose columns\slash entries provide the desired hyperplanes. These procedures solve problems of the form
\begin{equation}\label{eq:multiclass}
\left(\mW,\bb\right) := \underset{\mW \in \R^{K \times p}, \, \bb\in\R^K}{\text{arg min }} \sum_{i=1}^n L_{y_i}(\mW \xx_i + \bb) + \lambda R(\mW),
\end{equation}
where the $L_{y_i}$ promote correct classification of the training data and $R$ is a regularization penalty that controls model complexity.

The connection to embeddings occurs by noticing that the matrix $\mW$ implicitly embeds points en route to classifying them. If $r \leq \text{min}(K,p)$ is the rank of the coefficient matrix $\mW$, the latter may be decomposed into a product of $K \times r$ and $r \times p$ matrices,
\begin{equation}\label{eq:W_definition}
\mW = \mH\mPhi.
\end{equation}
The product $\mW\xx$ first maps $\xx$ into an $r$-dimensional subspace via the embedding $\mPhi$, whence the mapped vector $\mPhi\xx$ is compared to the rows $\hh_k$ of $\mH$. The ``winner-takes-all'' rule of \Eqnref{eq:multiclass} may be rewritten according to this decomposition as
\begin{equation}\label{eq:winner_takes_all_e}
y(\xx) = \underset{k\in\{1,\dots,K\}}{\text{arg max }} \hh_k^\top \left(\mPhi \xx \right)  + b_k.
\end{equation}
The embedding is nontrivial whenever $\mW$, hence $\mPhi$, has low rank.

Insight into why the linear map $\mPhi$ may serve as a useful embedding is given by comparing our multiclass framework with rank-restricted linear discriminant analysis (RR-LDA) \cite{friedman2001elements}. RR-LDA finds $\mPhi$ according to Gaussianity assumptions and simply labels each point $\xx \in \R^p$ according to its nearest class centroid $\mm_k = \sum_{\xx_i : y_i = k} \mPhi \xx_i$ once embedded, i.e.,\vspace{-2mm}
\[
y(\xx) =
\underset{k\in\{1,\dots,K\}}{\text{arg min }} \left\|\mm_k - \mPhi \xx\right\|_2^2 = \underset{k\in\{1,\dots,K\}}{\text{arg max }} \mm_k^\top \mPhi \xx -
\frac{1}{2}\left\|\mm_k\right\|_2^2,
\]
whenever class prior probabilities are equal. This classification rule is desirable for embeddings in that it encourages embedded points belonging to the same class to be close to one another (and separated from other points) by virtue of their proximity to the class centroid. It is also a special case of the ``winner-takes-all'' rule: the centroids play the role of $\hh_k$ and
$b_k = -\frac{1}{2}\|\mm_k\|_2^2$.
Whenever the RR-LDA solution is close to optimal for \Eqnref{eq:multiclass}---roughly speaking this is expected whenever RR-LDA provides good classification accuracy relative to learners optimizing the more general ``winner-takes-all'' rule---, the optimizer of \Eqnref{eq:multiclass} will behave similarly to the RR-LDA solution. In these cases optimizing \Eqnref{eq:multiclass} will result in an embedding that sensibly organizes the training data.\footnote{Further and more formal statements guaranteeing the quality of embeddings generated from \Eqnref{eq:multiclass} can be made by comparing Voronoi and convex polyhedral tesselations; these are omitted for brevity.}

Finally, it is important to highlight that the decomposition in \Eqnref{eq:W_definition} is not unique, as any $r \times r$ invertible matrix $\mV$ defines another valid decomposition via 
\begin{equation}
\mW = \mH\mPhi = (\mH \mV^{-1}) \left(\mV \mPhi\right) = {\mH'} {\mPhi'}.
\end{equation}
This indeterminacy is immaterial for classification using the learned hyperplanes, but it is of critical importance when the embedding is used to determine the relatedness of points potentially belonging to classes that do not appear in the training data. In this case we are forced to resort to a more generic comparison among points, e.g., Euclidean distance or cosine similarity, in the mapped space. Taking Euclidean distance as an example, simple counterexamples involving a diagonal $\mV$ show how $\left\|\mV\mPhi \uu - \mV\mPhi \vv \right\|_2$ can be made to essentially discard all information in $r-1$ of the dimensions. 

Selecting an optimal $\mV$ will be the topic of future work; for this work we focus on $\mPhi$ with orthonormal rows, e.g., as obtained from the economical singular value decomposition of $\mW$. This choice of $\mPhi$ is ``safe'' in that all dimensions contribute equally. Moreover, simple algebraic arguments show that any comparison based on inner products, such as Euclidean distance or cosine similarity, is invariant to the \emph{specific} choice of $\mPhi$ among the set of valid $\mPhi$ with orthonormal rows.

\xhdr{\Cl embedding}
We conclude this section by showing how to cast our crosslingual embedding problem as an instance of the above framework. We assume a universe of languages $\mathcal{L}$ from which we are given document samples $\mathcal{D}_l = \{d^l_1,\dots,d^l_{n_l}\}$ written in language $l\in\mathcal{L}$. Associated with each document $d_i^l$ is a class label $y^l_i \in \left\{1,\dots,K\right\}$ along with a \emph{language\hyp specific} feature representation $\xx_i^l \in \R^{p_l}$.
Our goal is to determine a linear mapping $\mPhi_l:\R^{p_l}\rightarrow \R^r$ for each language that embeds documents into a common ``semantic'' space, whereby documents with the same class label are closer to one another---irrespective of language---than they are to documents with different class labels.

The crosslingual embedding problem translates into our multiclass embedding framework by interpreting each $\xx_i^l$ as specifying the nonzero entries of a larger vector $\zz_i^l \in \R^p$ where $p = \sum_{l\in\mathcal{L}} p_l$. In words, each $\zz_i^l$ is a vector in a product space obtained by stacking the language\hyp specific feature representations, and it may only have nonzero entries in dimensions corresponding to the feature representation for language $l$. The $\zz_i^l$ may then directly be used in place of the $\xx_i^l$, and the discriminating hyperplane matrix $\mW$ along with $\mPhi$ will have a blockwise structure
\[\mW = \left[\mW_{l_1} \mid \mW_{l_2} \mid \dots \mid \mW_{l_{|\mathcal{L}|}}\right], \; \mPhi = \left[\mPhi_{l_1} \mid \mPhi_{l_2} \mid \dots \mid \mPhi_{l_{|\mathcal{L}|}}\right],\]
with each block pertaining to its respective language and acting only on the feature representation for that language. The desired embedding maps $\mPhi_l$ will simply be the blocks of $\mPhi$.

When using bag-of-word features, $\mPhi$ has one column per word in the vocabulary, so the columns of $\mPhi$ can be interpreted as word vectors.
Embedding a document entails multiplying its bag-of-word vector with $\mPhi$, which in turn corresponds to taking a weighted sum of the vectors corresponding to the words present in the document.

\xhdr{Training on Wikipedia}
In our concrete setting (\Secref{sec:eval}), documents are Wikipedia articles represented via bag-of-word features,
and an article's class label indicates the language\hyp independent concept the article is about, such that articles about the same concept in different languages will be close in the embedding space.
We emphasize that Wikipedia is only used for training; thereafter, arbitrary texts can be embedded.
(For more details, see \Secref{sec:experimental setup}.)

\subsection{\Cl reduced-rank ridge regression}
\label{sec:Cr5}

Any of the aforementioned multiclass classification algorithms described by \Eqnref{eq:multiclass} can be converted into a problem that modulates the dimensionality of the embedding space by imposing a rank constraint on $\mW$, i.e.,\vspace{-1mm}
\begin{equation}\label{eq:multiclass_rank}
\begin{aligned}
&\underset{\mW \in \R^{K \times p}, \, \bb\in\R^K}{\text{minimize }}& \sum_{i=1}^n L_{y_i}(\mW\xx_i + b) + \lambda R(\mW) \\
& \text{subject to }& \text{rank}(\mW) = r,
\end{aligned}
\end{equation}
or if we wish to maintain convexity, by adding a nuclear-norm penalty 
to the objective\vspace{-1mm}
\begin{equation}\label{eq:multiclass_nuclear}
\underset{\mW \in \R^{K \times p}, \, \bb\in\R^K}{\text{minimize }} \sum_{i=1}^n L_{y_i}(\mW\xx_i + \bb) + \lambda R(\mW) + \alpha \|\mW\|_*.
\end{equation}

The optimization problem pertaining to one-\vs-all regularized least squares classification
(also known as \textit{ridge regression,} hence the name \textit{Cr5,} for \textit{\textbf{Cr}osslingual \textbf{r}educed\hyp \textbf{r}ank \textbf{r}idge \textbf{r}egression})
is particularly amenable to these additional constraints; it can be solved \emph{exactly} with the rank condition in \Eqnref{eq:multiclass_rank}, and it admits a massively scalable optimization procedure that easily extends to millions of classes. In this case, \Eqnref{eq:multiclass_rank} becomes
\begin{equation}\label{eq:rlsc1}
\begin{aligned}
&\underset{\mW \in \R^{K \times p}, \, \bb\in\R^K}{\text{minimize }}& \frac{1}{2}\big\|\mY - \mX\mW^\top - \mathbf{1}\bb^\top\big\|_F^2 + \frac{\lambda}{2} \|\mW\|_F^2 \\
& \text{subject to }& \text{rank}(\mW) = r,
\end{aligned}
\end{equation}
where $\mY \in \R^{n \times K}$ is the one-hot encoding of the class label for each of the $n$ training points (documents), i.e., $Y_{iy_i}$ are its only nonzero entries, and $\mX \in \R^{n \times p}$ stores the training points as its rows. We will show that \Eqnref{eq:rlsc1} is equivalent to a singular value decomposition problem by first eliminating the offsets $\bb$ and then manipulating the resulting quadratic problem into appropriate form. These types of derivations appear throughout the statistics literature and are used to show that the RR-LDA solution can be obtained from reduced\hyp rank ridge regression \cite{hastie1994flexible}.

First, the optimality conditions for $\bb$ imply
$$\bb = \frac{1}{n} \left(\mY^\top - \mW\mX^\top \right)\mathbf{1}.$$
Plugging this into \Eqnref{eq:rlsc1} allows us to reduce the problem to
\begin{equation}\label{eq:rlsc2}
\begin{aligned}
&\underset{\mW \in \R^{K \times p}}{\text{minimize }}& \frac{1}{2} \text{trace}\left[\mW\big(\hat{\mX}^\top\hat{\mX} + \lambda \mI\big)\mW^\top\right] -\text{trace}\left[\hat{\mY}^\top\hat{\mX}\mW^\top\right]\\
& \text{subject to }& \text{rank}(\mW) = r,
\end{aligned}
\end{equation}
where the notation $\hat{\mM} := \mM - \frac{1}{n}\mathbf{1}\mathbf{1}^\top \mM$ denotes the column-wise mean\hyp centered version of any $n \times t$ matrix $\mM$. The matrix
\[\hat{\mX}^\top \hat{\mX} + \lambda \mI = \mL\mL^\top\]
admits a Cholesky decomposition because it is symmetric positive definite for $\lambda > 0$, so we define a new optimization variable $\mZ := \mW\mL$, which allows us to rewrite \Eqnref{eq:rlsc2} as
\begin{equation}\label{eq:rlsc3}
\begin{aligned}
&\underset{\mZ \in \R^{K \times p}}{\text{minimize }}& \frac{1}{2} \Big\|\hat{\mY}^\top\hat{\mX}\big(\mL^{-1}\big)^\top - \mZ \Big\|_F^2\\
& \text{subject to }& \text{rank}(\mZ) = r
\end{aligned}
\end{equation}
after completing the square. Note that $\text{rank}(\mW)=\text{rank}(\mZ)$ because~$\mL$ is full rank. It is well known that \Eqnref{eq:rlsc3} characterizes the singular value decomposition of $\hat{\mY}^\top\hat{\mX}\left(\mL^{-1}\right)^\top$ and can therefore be solved by a variety of methods for computing this decomposition.

\xhdr{Iterative solution}
While we have used the Cholesky decomposition to show that the problem in \Eqnref{eq:rlsc1} is equivalent to a singular value decomposition, it is not necessary to compute this Cholesky decomposition. Indeed, we are particularly interested in situations where $n,K,$ and $p$ are so large that it is impossible to directly compute a decomposition of any of the matrices involved, so we must instead rely on iterative methods. We now show how to compute all necessary quantities using such iterative methods.

We begin by observing that, if
\begin{equation}\label{eq:svd_rlsc}
\hat{\mY}^\top\hat{\mX}\left(\mL^{-1}\right)^\top = \mP\mSigma \mV^\top    
\end{equation}
is the singular value decomposition of the argument in \Eqnref{eq:rlsc3}, then
\begin{equation}\label{eq:eig_matrix}
\hat{\mY}^\top\hat{\mX}\left(\hat{\mX}^\top\hat{\mX} + \lambda \mI\right)^{-1}\hat{\mX}^\top\hat{\mY} = \mP\mSigma^2\mP^\top    
\end{equation}
provides $\mP, \mSigma$ from its eigenvalue decomposition. This $K\times K$ matrix does not involve $\mL$ and instead relies on an inverse matrix. Iterative eigenvalue solvers rely on computing matrix--vector products of the form
$\vv = \hat{\mY}^\top\hat{\mX}\left(\hat{\mX}^\top\hat{\mX} + \lambda \mI\right)^{-1}\hat{\mX}^\top\hat{\mY}\uu$,
so $\vv$ can be obtained by using an iterative equation solver, such as a conjugate gradient method, to solve
\begin{equation}
\label{eq:conj grad}
\left(\hat{\mX}^\top\hat{\mX} + \lambda \mI \right)\xx=\hat{\mX}^\top\hat{\mY}\uu
\end{equation}
and to then compute the product $\vv=\hat{\mY}^\top\hat{\mX}\xx$.

Suppose now that we have computed the $r$ largest eigenvectors in \Eqnref{eq:eig_matrix}, which we denote by $\mP_{[r]} \in \R^{K \times r}$.
The fastest way to
decompose $\mW$ into
separating hyperplanes and
an embedding map as specified in \Eqnref{eq:W_definition} is via
\begin{equation}\label{eq:answer}
\begin{aligned}
&\mH_\star = \mP_{[r]} \\
&\mPhi_\star = \mP_{[r]}^\top\hat{\mY}^\top\hat{\mX} \left(\hat{\mX}^\top\hat{\mX} + \lambda \mI\right)^{-1},
\end{aligned}
\end{equation}
where we may again use an iterative equation solver to compute~$\mPhi_\star$. 

The matrices $\mH_\star,\mPhi_\star$ also allow us to compute the singular value decomposition of $\mW$.
This is accomplished by observing that, if
$\mPhi_\star\mPhi_\star^\top=\mQ \mLambda \mQ^\top$
is an eigenvalue decomposition of the $r\times r$ matrix $\mPhi_\star\mPhi_\star^\top$, then
\begin{equation}
    \mW\mW^\top=\mH_\star\mPhi_\star\mPhi_\star^\top \mH_\star^\top=\left(\mH_\star\mQ\right) \mLambda \left(\mH_\star\mQ\right)^\top
\end{equation}
provides an eigenvalue decomposition of $\mW\mW^\top$ since
$\mQ^\top\mH_\star^\top\mH_\star\mQ = \mI$.
This eigenvalue decomposition is fast to compute directly whenever~$r$ is no more than several tens of thousands, and we can extract the singular value decomposition
$\mW= \left(\mH_\star\mQ\right)\mLambda^{\frac{1}{2}} \mT^\top$
via
$\mT^\top = \mLambda^{-\frac{1}{2}}\mQ^\top\mPhi_\star$.

\section{Evaluation}
\label{sec:eval}

This section showcases the performance of the \name embedding method of \Secref{sec:Cr5} on retrieval tasks involving text at various levels of granularity (from documents to sentences to words) and provides a comparison with the current state-of-the-art models.
As \name has been primarily designed for handling entire documents, our first and main evaluation considers \cl document retrieval (\Secref{sec:doc retrieval}). 
In order to determine how well our method works on shorter pieces of text, we also test it on \cl sentence (\Secref{sec:sentence retrieval}) and word (\Secref{sec:word retrieval}) retrieval tasks.
Before we present results, we first describe our experimental setup (\Secref{sec:experimental setup}).


\subsection{Experimental setup}
\label{sec:experimental setup}

\xhdr{Training data: multilingual Wikipedia}
As described in \Secref{sec:rank-reduced classification}, our method leverages a corpus of class\hyp labeled documents and strives to place documents from the same class close to one another in the embedding space.
We use Wikipedia as our document collection for training, since a large fraction of its articles exist in multiple languages.
Additionally, articles are aligned across languages: each article is attributed to the language\hyp independent concept it is about (\eg, both English \cpt{beer} and Italian \cpt{birra} are attributed to the concept Q44), and we use these concepts as our class labels.
Consequently, the number of classes equals the number of unique Wikipedia concepts across all languages (millions),
while the number of members per class is upper\hyp bounded by the number of languages considered, as each language has at most one article about each language\hyp independent concept.


\xhdr{Languages}
While \name can in principle handle any number of languages,
for reasons of clarity, we focus on 4 languages here: English (en), Italian (it), Danish (da), and Vietnamese (vi).
English and Italian were chosen because the pair has often been used in the prior literature \cite{dinu2014improving, smith2017offline, conneau2017word} and because those languages come with large Wikipedia versions (5.7m and 1.5m articles, respectively, at the time of writing).
Danish was chosen because it has a much smaller Wikipedia and thus training set (239k articles),
and Vietnamese (1.1m articles), because, due to Vietnam's geographical and cultural distance from Europe, its Wikipedia version has much less overlap with European languages than those do among each other, which will demonstrate that \name also works on distant language pairs.

\xhdr{Retrieval}
Our evaluation consists in \cl text retrieval tasks, where we consider as texts entire documents, sentences, and single words.
Given a pair of a \textit{query language} $l_q$ and a \textit{target language} $l_t$, as well as a \textit{query text} written in the query language $l_q$, the objective is to rank all \textit{candidate texts} in the target language $l_t$ such that the  $l_t$-text corresponding to the  $l_q$-query is top\hyp ranked, where the ranking is done in decreasing order of similarity between the query and the candidates, with similarity measured in the common embedding space.
The most straightforward similarity measure for ranking is the cosine, which we therefore use as our main measure.
Prior work has also experimented with different, more complex measures; \eg, Conneau et al. \cite{conneau2017word} use \textit{cross\hyp domain similarity local scaling} (CSLS), which is proposed in order to mitigate the so-called ``hubness problem'' \cite{radovanovic2010hubs, dinu2014improving}, which causes some central points to be close to nearly all other points.
Therefore, in addition to cosine similarity, we also consider CSLS and report results for both measures.
Following prior work \cite{conneau2017word, smith2017offline}, results are reported in terms of what is there called \textit{precision at~$k$} (P@$k$) for $k = 1,5,10$, defined as the fraction of queries whose correct equivalent is found among the $k$ top\hyp ranked points from the target language.

\xhdr{Baseline}
In our experiments, we consider the best\hyp performing, unsupervised model of Conneau et al. \cite{conneau2017word} (\baseline) as our main baseline, as it has been shown to outperform other methods by a wide margin (\cf\ \Secref{sec:relwork}) on crosslingual document retrieval \cite{litschko2018unsupervised}, which is our primary focus, as well as on word and sentence retrieval.
To represent longer texts using \baseline word embeddings, we use the authors' code, which weights individual words by their inverse document frequency (IDF) and averages their vectors.

\xhdr{Data preprocessing}
For both training and testing, we represent input texts ($\xx_i$ in \Secref{sec:rank-reduced classification}) as TF-IDF--weighted bag-of-word vectors, where IDF weights are computed on the training set only. 
 In order to avoid noise arising from very short and very long articles, we
(1)~filter the training corpus down to documents containing between 50 and 1,000 unique 
words (which covers the bulk of Wikipedia articles), and
(2)~normalize input bag-of-word vectors to $L_2$\hyp unit-length ($L_1$\hyp normalization gave slightly worse results in pilot runs).
As a further preprocessing step, words are lower\hyp cased, and the vocabulary is restricted to the 200k most frequent words, after discarding words that appear in fewer than 3 documents.

\xhdr{Hyperparameters}
While \name offers several hyperparameters, our experiments showed that most of them can be fixed to globally optimal values,
including tolerance thresholds $\epsilon$ and maximum iteration numbers $T$ for the conjugate\hyp gradient ($\epsilon=0.01, T=500$) and eigenvalue\hyp decomposition ($\epsilon=0.1, T=250$) routines.
We found that performance increases with the dimensionality $r$ of the embedding space and fix $r=300$ to trade off performance \vs computation time.
Hence, the only parameter that remains to be cross\hyp validated is the regularization parameter $\lambda$.

\xhdr{Code and data}
Our code and pretrained embeddings for 28 languages are available at
 \url{https://github.com/epfl-dlab/Cr5}.

\subsection{Document retrieval}
\label{sec:doc retrieval}

As \name was conceived for document embedding, our main evaluation task considers \cl document retrieval.

\begin{figure}[t]
\subfigure[Joint training]{\label{fig:venn_joint}\includegraphics[width=0.32\columnwidth]{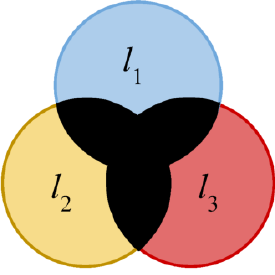}}
\subfigure[Pairwise training]{\label{fig:venn_pairs}\includegraphics[width=0.32\columnwidth]{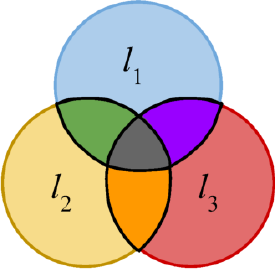}}
\subfigure[Transitive training]{\label{fig:venn_trans}\includegraphics[width=0.32\columnwidth]{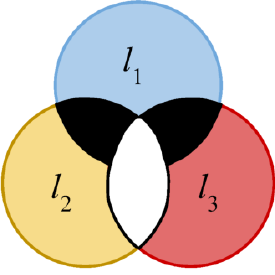}}
\vspace{-4mm}
\caption{
Three ways of training \name; details in \Secref{sec:doc retrieval}.
}
\label{fig:venn}
\vspace{-4mm}
\end{figure}

\xhdr{Train\slash test splits}
To explain how we split the data into training and testing sets, consider \Figref{fig:venn}, which contains Venn diagrams of Wikipedia in three languages $l_1, l_2, l_3$ and where, in a slight abuse of notation, we also let $l_i$ represent the set of concepts for which there is a Wikipedia article in language $l_i$.
For each language pair $(l_i, l_j)$ on which we want to evaluate, we first take out 2k concepts from $l_i \cap l_j$: 1k to be used as queries for testing, and 1k, as queries for cross\hyp validation.
Then, for each language $l_i$, we take out 200k concepts (or fewer if there are not as many) that will serve as retrieval candidates when $l_i$ is used as the target language
(\ie, a random ranking would yield a very low precision at 1/5/10 of 0.0005\%\slash 0.0025\%\slash 0.005\%).
Finally, the remaining documents from $l_i \cap l_j$ are used as the training set for the pair $(l_i, l_j)$.

We consider three training settings:
(1)~\textit{Joint training} (\Figref{fig:venn_joint}) uses the union of pairwise intersections as training data and fits a single model to be used to embed documents from any of the languages considered.
(2)~\textit{Pairwise training} (\Figref{fig:venn_pairs}) considers each pairwise intersection individually and fits a separate model to be used for the respective pair only.
(3)~\textit{Transitive training} (\Figref{fig:venn_trans}) simulates a scenario where two languages have no common concepts, but each share some concepts with a third language.
Next, we present results for each of these settings in turn.


\begin{table*}[t]
\begin{subtable}
\centering
\caption{
Precision at $k$
on document retrieval.
One row per language pair ($l_1$, $l_2$; codes resolved in \Secref{sec:experimental setup}); each pair evaluated in both directions (query in $l_1$, candidates in $l_2$, and vice versa) and using both similarity measures (cosine, CSLS; \cf\ \Secref{sec:experimental setup}).
}
\label{tab:mastertable}
\begingroup
\setlength{\tabcolsep}{4pt} 
\renewcommand{\arraystretch}{0.92}

\begin{tabular}{ccc||rrr|rrr||rrr|rrr||r}
& \multirow{2}{*}{$l_1$} & \multirow{2}{*}{$l_2$}   & \multicolumn{3}{c|}{Query in $l_1$ (cosine)}  & \multicolumn{3}{c||}{Query in $l_2$ (cosine)} & \multicolumn{3}{c|}{Query in $l_1$ (CSLS)} & \multicolumn{3}{c||}{Query in $l_2$ (CSLS)} & \multicolumn{1}{c}{\# docs in} \\
\cline{4-15}
 &  & & P@1  & P@5  & P@10 & P@1  & P@5  & P@10 & P@1   & P@5  & P@10 & P@1          & P@5         & \multicolumn{1}{c||}{P@10} & \multicolumn{1}{c}{intersection}       \\
\hline
\multicolumn{16}{c}{\vspace{-2mm}}\\
\multicolumn{16}{c}{\textbf{(a) \baseline \cite{conneau2017word}}}\\
\hline
$\star$ & da & en & 37.7 & 54.8 & 61.1 & 27.6 & 41.0 & 47.1 & 46.4  & 63.2 & 70.0 & 35.8         & 52.5        & 59.8 &        \\
 & da & it & 22.2 & 39.2 & 45.8 & 16.2 & 30.6 & 38.0 & 28.2  & 46.4 & 53.0 & 24.5         & 42.5        & 51.9 &        \\
$\bullet$ & da & vi & 5.3  & 11.8 & 17.3 & 5.8  & 13.8 & 18.5 & 8.1   & 18.9 & 26.1 & 11.9         & 24.1        & 30.2 &        \\
$\circ$ & en & it & 33.3 & 46.8 & 52.8 & 38.9 & 54.9 & 62.6 & 40.0  & 57.3 & 63.1 & 49.2         & 62.9        & 69.0 &        \\
 & en & vi & 7.5  & 14.3 & 18.8 & 11.1 & 23.4 & 28.6 & 11.7  & 25.0 & 31.7 & 23.6         & 40.6        & 47.9 &        \\
 & it & vi & 4.2  & 9.1  & 12.6 & 7.2  & 14.6 & 19.0 & 7.4   & 15.9 & 22.3 & 13.2         & 23.8        & 30.8 &        \\
\hline
\multicolumn{16}{c}{\vspace{-2mm}}\\
\multicolumn{16}{c}{\textbf{(b) \name, joint training (\Figref{fig:venn_joint})}}\\
\hline
$\star$ & da & en & 62.6 & 76.4 & 80.4 & 53.7 & 74.5 & 81.9 & 68.4  & 83.5 & 86.9 & 50.9         & 74.2        & 81.0 &     79k   \\
 & da & it & 41.2 & 63.2 & 70.9 & 41.9 & 66.4 & 74.1 & 48.2  & 68.9 & 76.9 & 38.6         & 64.3        & 73.7 &    59k    \\
$\bullet$ & da & vi & 33.7 & 54.8 & 62.8 & 34.6 & 59.0 & 67.9 & 36.4  & 60.2 & 67.3 & 36.3         & 59.8        & 70.4 &   25k     \\
$\circ$ & en & it & 70.0 & 85.0 & 88.9 & 78.2 & 89.2 & 91.1 & 70.5  & 85.4 & 90.1 & 79.0         & 90.1        & 92.5 &     489k   \\
 & en & vi & 50.4 & 68.4 & 75.9 & 63.0 & 79.2 & 83.4 & 47.3  & 70.1 & 78.5 & 68.7         & 82.8        & 87.2 &    98k    \\
 & it & vi & 38.9 & 61.9 & 69.4 & 45.5 & 66.1 & 73.3 & 35.9  & 61.9 & 70.5 & 48.3         & 70.2        & 77.1 &    62k    \\
\hline
\multicolumn{16}{c}{\vspace{-2mm}}\\
\multicolumn{16}{c}{\textbf{(c) \name, pairwise training (\Figref{fig:venn_pairs})}}\\
\hline
$\star$ & da & en & 69.2        & 82.7        & 86.0       & 58.9        & 76.4        & 82.2        & 70.1         & 84.0        & 87.7        & 57.3         & 75.7         & 82.7 &    74k    \\
$\bullet$ & da & vi & 32.0        & 55.9        & 63.5       & 33.9        & 58.2        & 66.1        & 35.6         & 60.4        & 68.9        & 33.8         & 56.3         & 66.4 &    22k    \\
$\circ$ & en & it & 74.6        & 86.4        & 88.9       & 80.0        & 89.8        & 91.5        & 71.9         & 85.8        & 90.3        & 79.0         & 90.2         & 91.9 &    483k    \\
\hline
\multicolumn{16}{c}{\vspace{-2mm}}\\
\multicolumn{16}{c}{\textbf{(d) \name, pairwise training (low resources; \Figref{fig:venn_pairs})}}\\
\hline
$\bullet$ & da & vi & 29.5        & 50.2        & 58.1       & 24.5        & 48.7        & 57.2        & 24.3         & 44.4        & 53.3        & 24.8         & 47.9         & 55.9 &    10k    \\
\hline
\multicolumn{16}{c}{\vspace{-2mm}}\\
\multicolumn{16}{c}{\textbf{(e) \name, transitive training  (\Figref{fig:venn_trans})}}\\
\hline
$\bullet$ & da & vi & 21.6        & 39.8        & 47.4       & 25.1        & 44.7        & 53.5        & 27.8         & 49.3        & 60.0        & 27.1         & 49.6         & 59.1 &  0 \\
\hline
\end{tabular}
\end{subtable}
\endgroup


\end{table*}

\xhdr{Joint training (\Figref{fig:venn_joint})}
We start by training a multilingual model on all 4 languages (Danish, English, Italian, Vietnamese) simultaneously and testing it on all 12 directed pairs.
To preclude overfitting, any one of the 2k concepts sampled as queries for testing and cross\hyp validation for any language pair is also excluded from all other pairwise intersections when building the training sets.

\begin{figure}[t]
\includegraphics[width=1.1in]{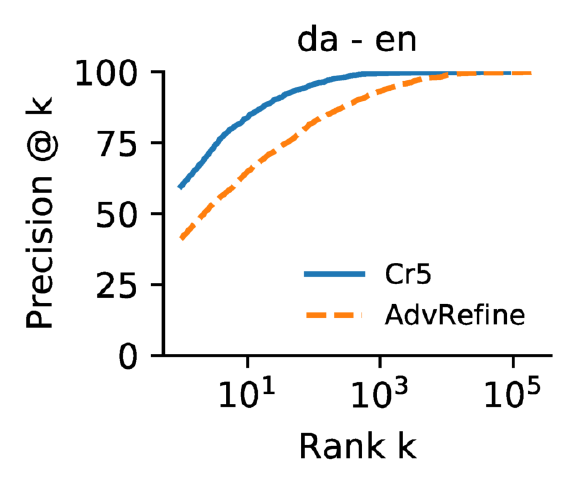}
\hspace{-1mm}
\includegraphics[width=1.1in]{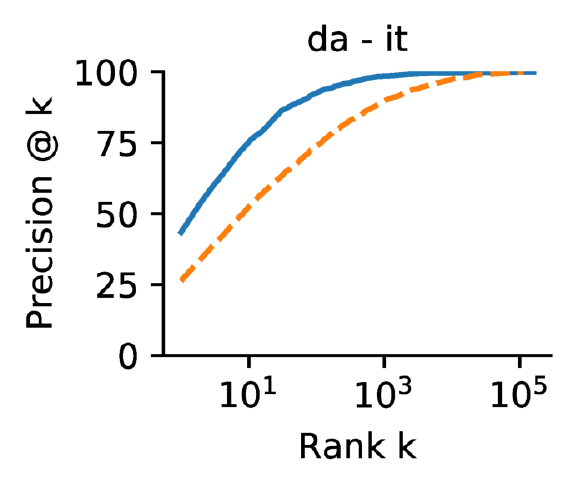}
\hspace{-1mm}
\includegraphics[width=1.1in]{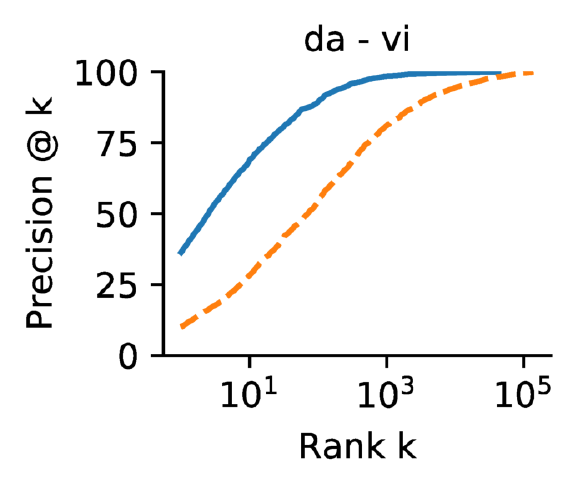}

\includegraphics[width=1.1in]{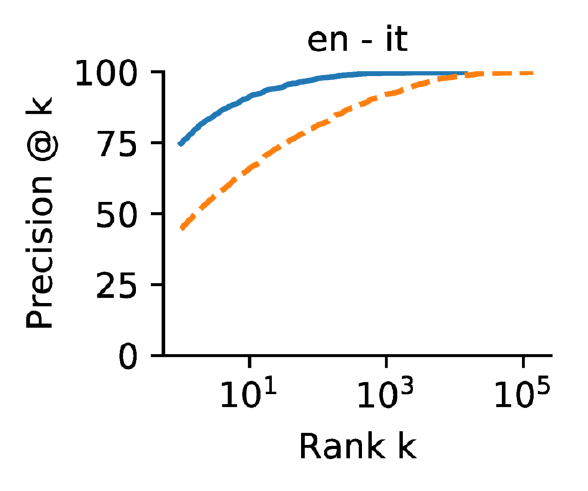}
\hspace{-1mm}
\includegraphics[width=1.1in]{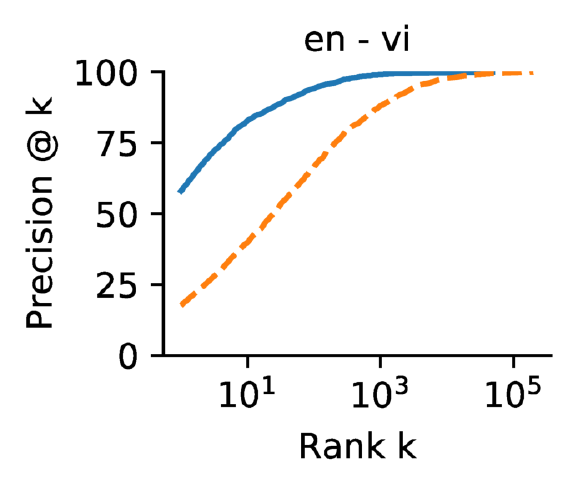}
\hspace{-1mm}
\includegraphics[width=1.1in]{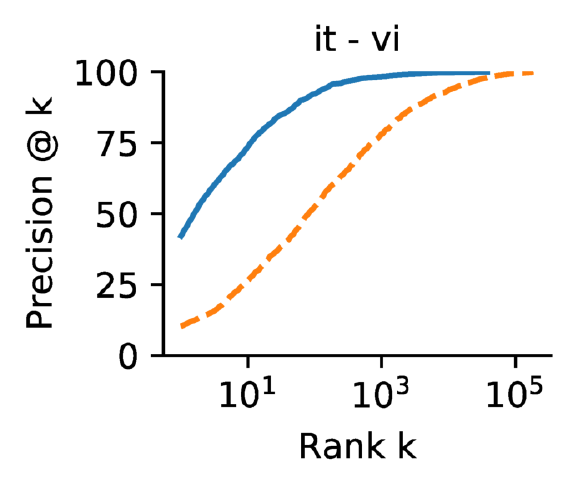}

\vspace{-4mm}
\caption{
Precision at $k$ on document retrieval, using CSLS similarity measure.
\name trained jointly on Danish (da), English (en), Italian (it), Vietnamese (vi) (\cf\ \Tabref{tab:mastertable}(b)).
Each pair was evaluated in both directions, average is plotted.
}
\label{fig:cdf}
\vspace{-4mm}
\end{figure}

\Tabref{tab:mastertable}(b) summarizes the performance of our method in terms of precision at 1, 5, and 10, for both similarity measures (cosine and CSLS, \cf \Secref{sec:experimental setup}).
(Since CSLS is better everywhere for \baseline, and nearly everywhere for \name, we focus on this measure in our discussion.)
First, we note that the performance of \name is overall high in absolute terms:
the smallest P@1 (P@10) is 35.9\% (67.3\%), \ie, for all pairs, the correct target document is ranked first for at least one third of all queries, and among the top 10 for at least two thirds of them.
Moreover, comparing our results to \Tabref{tab:mastertable}(a), we see that we clearly outperform \baseline \cite{conneau2017word}, the previous state-of-the-art method according to Litschko et al.\ \cite{litschko2018unsupervised}, on all language pairs;
\eg, \baseline achieves its highest P@1 of 49.2\% when retrieving English documents for Italian queries, while \name achieves 79.0\% in this setting.
The gains are especially high for pairs involving low\hyp resource languages, such as Danish and Vietnamese, where we increase P@1 from 8.1\% to 36.4\%.
These outcomes are echoed visually by \Figref{fig:cdf}, which
plots precision at $k$ as a function of the rank $k$
and where \name's curve lies far above \baseline's curve for all pairs.

To illustrate our results, we provide two examples in \Tabref{tbl:doc retrieval example}.
On the left, we list the Danish articles (titles translated to English) that are closest to the Vietnamese query \cpt{Supercomputer} (\cpt{Si\^eu m\'ay t\'inh}) in the embedding space;
on the right, the English articles that are closest to the Danish query \cpt{Traditional heavy metal} (\cpt{Klassisk metal}).
The second example showcases that, even when we do not retrieve the true target at rank 1, we still retrieve semantically close concepts.
In other words, our embedding space captures semantic similarity across languages.

\begin{table}[t]
\centering
\caption{
Examples of \cl document retrieval on Wikipedia.
Article names given in their English versions.
}
\vspace{-2mm}
\label{tbl:doc retrieval example}
\begingroup
\renewcommand{\arraystretch}{0.92}
{\small
\begin{tabular}{l}
\textbf{Query language: Vietnamese}\\
\textbf{Target language: Danish}\\
\textbf{Query: \cpt{Supercomputer}}\\
\hline
\textbf{\cpt{Supercomputer}} \\
\cpt{Central processing unit} \\
\cpt{Tianhe-I} \\
\cpt{IBM Personal Computer} \\
\cpt{Mainframe computer} \\
\cpt{Computer} \\
\cpt{IBM} \\
\cpt{Unified EFI Forum} \\
\cpt{Moblin} \\
\cpt{Barebone computer}
\end{tabular}
\begin{tabular}{l}
\textbf{Query language: Danish}\\
\textbf{Target language: English}\\
\textbf{Query: \cpt{Trad.\ heavy metal}}\\
\hline
\cpt{South African heavy metal} \\
\cpt{New wave of Amer.\ heavy metal} \\
\cpt{Cyber metal} \\
\cpt{List of folk metal bands} \\
\textbf{\cpt{Traditional heavy metal}} \\
\cpt{Season of Mist} \\
\cpt{Blackgaze} \\
\cpt{Wayd} \\
\cpt{List of speed metal bands} \\
\cpt{Worship Him}
\end{tabular}
}
\endgroup
\vspace{-2mm}
\end{table}

\xhdr{Pairwise training (\Figref{fig:venn_pairs})}
In the above experiments, we trained a single model to be used for embedding documents from any language.
If, however, a downstream application involves only documents from a specific language pair (\eg, when documents in only one fixed language $l_1$ are to be retrieved for queries in another fixed language $l_2$), we may also train a separate model for each pair.
In order to investigate how this affects performance, we train separate models for three pairs, the first coupling two high\hyp resource languages (English and Italian), the second, two low\hyp resource languages (Danish and Vietnamese), and the third, a low- with a high\hyp resource language (Danish and English).
The results are given in \Tabref{tab:mastertable}(c).
Comparing them to the results from joint training (\Tabref{tab:mastertable}(b)), we observe that pairwise training slightly improves performance for pairs involving at least one high\hyp resource language, whereas joint training tends to benefit the low\hyp resource pair Danish\slash Vietnamese.
We argue that this is because, when a pair has a small intersection, it can still benefit from transitivity:
while many concepts may not be shared by Danish and Vietnamese, there may be a large set of concepts that is shared by Danish and English, and another, disjoint set that is shared by Vietnamese and English (\Figref{fig:venn_trans}), such that information can effectively flow between Danish and Vietnamese via English (we confirm this intuition in a separate experiment below).
Additionally, vocabularies are larger when using the union of all pairwise intersections for training, rather than individual pairwise intersections, so the model is better equipped for embedding unseen queries and candidates during testing.


We see that, even for the low\hyp resource pair Danish\slash Vietnamese, performance is acceptable under pairwise training;
\eg, the correct target is ranked first for one third of all queries, and among the top~10 for two thirds of them.
In order to emphasize this point, we further decrease the intersection artificially, reducing it from 22k to 10k concepts.
The results in \Tabref{tab:mastertable}(d) show that, while, as expected, performance drops significantly, the absolute numbers are still acceptable;
\eg, the correct target is ranked first for one quarter of all queries, and among the top 10 for over half of them.



\xhdr{Transitive training (\Figref{fig:venn_trans})}
To further investigate the above remark about transitivity, we now simulate a scenario where two languages (Danish and Vietnamese) share no concepts at all, but each have overlapping concepts with a third language (English).
In particular, we exclude the concepts that are described in both Danish and Vietnamese from training and consider only the documents that are described either in Danish and English or in Vietnamese and English, but not in both.
\Tabref{tab:mastertable}(e) shows that, although the performance for Danish\slash Vietnamese drops by around 10\% percent compared to joint training (\Tabref{tab:mastertable}(a)), it is still higher than that of \baseline \cite{conneau2017word,litschko2018unsupervised} by a factor of 2 to 3.5.


\xhdr{Summary}
We may thus summarize our performance on document retrieval by stating that
(1)~\name outperforms the previous state-of-the-art method \baseline \cite{conneau2017word,litschko2018unsupervised} by a wide margin, especially for low\hyp resource languages, and that
(2)~pairwise training works better for most language pairs, with the exception that
(3)~joint training works slightly better for low\hyp resource languages, partly because it enables transitive information flow between languages.

\subsection{Sentence retrieval}
\label{sec:sentence retrieval}
\name has been designed for embedding documents, and the previous section showed that it performs very well on document retrieval.
We now explore whether the method can also be used on shorter units of text, such as sentences and individual words.
For \cl sentence retrieval, we use the \textit{Europarl} corpus \cite{koehn2005europarl}, which contains millions of sentences from European Parliament debates translated into 11 European languages each and aligned sentence by sentence.
Following prior work, we focus on the pair English\slash Italian.
We use 2k sentences as queries, 200k sentences (including the 2k queries) as retrieval candidates, and 300k separate sentences for computing IDF weights and cross\hyp validating the regularization parameter.
This setting has been used in previous work \cite{conneau2017word,smith2017offline} and thus facilitates direct comparison to related methods.

For the evaluation, we first map all query and candidate sentences to the embedding space using the pairwise English\slash Italian model trained on Wikipedia (\Secref{sec:doc retrieval}; performance is similar for the joint model).
Then, given a query sentence, we aim at retrieving its correct translation in the target language.
The results (\Tabref{tab:sentenceTranslation}) show that for about half of all queries, we retrieve the correct translation at rank 1, and for two thirds, within the top 10.
Comparing to prior methods, we perform consistently better than Dinu et al.\ \cite{dinu2014improving} and Mikolov et al.\ \cite{mikolov2013distributed},
consistently worse 
than Procrustes (CSLS) \cite{conneau2017word} and \baseline (CSLS) \cite{conneau2017word},
and partly better, partly worse, than Smith et al.\ \cite{smith2017offline} and Procrustes (cosine)~\cite{conneau2017word}.
As our embedding is trained at a vastly different level of text granularity (documents rather than sentences) and on a vastly different type of corpus (Wikipedia rather than parliament debates), these results demonstrate that \name generalizes well across text lengths and corpora.




\begin{table}[t]
\caption[Sentence translation]{Precision at $k$ on sentence retrieval.%
\footnote{\label{test}}
(``\baseline (cosine)'' not reported by authors \cite{conneau2017word}.)
}
\vspace{-3mm}
\label{tab:sentenceTranslation}
\resizebox{\columnwidth}{!}{
\begingroup
\renewcommand{\arraystretch}{0.92}
\begin{tabular}{l|ccc|ccc}
                       & \multicolumn{3}{c|}{Query in Italian} & \multicolumn{3}{c}{Query in English} \\ \hline
                       & P@1         & P@5         & P@10        & P@1         & P@5         & P@10       \\ \hline
Mikolov et al. \cite{mikolov2013exploiting} & 10.5        & 18.7        & 22.8        & 12.0        & 22.1        & 26.7       \\
Dinu et al. \cite{dinu2014improving}    & 45.3        & 72.4        & 80.7        & 48.9        & 71.3        & 78.3       \\
Smith et al. \cite{smith2017offline}    & 54.6        & 72.7        & 78.2        & 42.9        & 62.2        & 69.2       \\
Procrustes (cosine)  \cite{conneau2017word}      & 42.6        & 54.7        & 59.0        & 53.5        & 65.5        & 69.5       \\
Procrustes (CSLS) \cite{conneau2017word}    & 66.1        & 77.1        & 80.7        & 69.5        & 79.6        & 83.5       \\
\baseline (CSLS) \cite{conneau2017word}            & 65.9        & 79.7        & 83.1        & 69.0        & 79.7        & 83.1       \\ \hline
\name (cosine)              & 42.6        & 55.1        & 59.7        & 45.8        & 56.6        & 61.6       \\
\name (CSLS)             & 49.7        & 62.0        & 66.7        & 50.4        & 62.7        & 66.8      \\
\hline
\end{tabular}
\endgroup
}
\end{table}
\footnotetext{\label{test}Results for baselines are from Conneau et al. \cite{conneau2017word}.}

\subsection{Word retrieval}
\label{sec:word retrieval}
In our final evaluation, we go one step further and test whether our embedding also works at an even finer level of granularity, on a word (rather than document or sentence) retrieval task, where the goal is to retrieve the exact translation of a word from the query language in the target language.
For a direct comparison with previous approaches, we report results on the bilingual English\slash Italian dictionary released by Dinu et al.\ \cite{dinu2014improving}, using 1,500 query words and 200k retrieval candidates.
The dictionary contains an additional set meant to be used for training, on which we cross\hyp validate the regularization parameter.
As for sentence retrieval (\Secref{sec:sentence retrieval}), we train our embedding at the document level on Wikipedia in a pairwise English\slash Italian setup.
As word embeddings, we use the columns of matrix $\mPhi$ (\Eqnref{eq:W_definition}), which maps from bag-of-words space to embedding space, such that each of its columns corresponds to one word of the vocabulary and may thus be interpreted as a word vector.

The results are summarized in Table \ref{tab:wordTranslation}.
Focusing on P@1, we observe that, for over 40\% of words, we retrieve the true translation at rank 1, thereby outperforming all methods except Procrustes \cite{conneau2017word} and \baseline \cite{conneau2017word}.
Interestingly, however, the increase from P@1 to P@5 and P@10 is less pronounced here than for sentence retrieval, such that several methods achieve higher performance than \name in terms of P@5 and P@10.
Nonetheless, given that the competing methods were trained specifically for embedding words, whereas ours was trained for embedding documents, we are pleased with this competitive performance at the word level.

\begin{table}[t]
\caption{Precision at $k$ on word retrieval.%
\textsuperscript{\ref{test}}
(``Procrustes (cosine)'' and ``\baseline (cosine)'' omitted by authors \cite{conneau2017word}.)
}
\vspace{-3mm}
\label{tab:wordTranslation}
\resizebox{\columnwidth}{!}{
\begingroup
\renewcommand{\arraystretch}{0.92}
\begin{tabular}{l|ccc|ccc}
                         & \multicolumn{3}{c|}{Query in Italian} & \multicolumn{3}{c}{Query in English} \\ \hline
                         & P@1         & P@5         & P@10       & P@1         & P@5         & P@10        \\ \hline
Mikolov et al. \cite{mikolov2013exploiting}   & 33.8        & 48.3        & 53.9       & 24.9        & 41.0        & 47.4        \\
Dinu et al. \cite{dinu2014improving}       & 38.5        & 56.4        & 63.9       & 24.6        & 45.4        & 54.1        \\
Faruqui and Dyer \cite{faruqui2014improving}                      & 36.1        & 52.7        & 58.1       & 31.0        & 49.9        & 57.0        \\
Artetxe et al. \cite{artetxe2017learning}    & 39.7        & 54.7        & 60.5       & 33.8        & 52.4        & 59.1        \\
Smith et al. \cite{smith2017offline}   & 43.1        & 60.7        & 66.4       & 38.0        & 58.5        & 63.6        \\
Procrustes (CSLS) \cite{conneau2017word}  & 63.7        & 78.6        & 81.1       & 56.3        & 76.2        & 80.6        \\
\baseline (CSLS) \cite{conneau2017word}         & 66.2        & 80.4        & 83.4       & 58.7        & 76.5        & 80.9        \\ \hline
\name (cosine)                 & 41.0        & 51.8        & 55.0         & 38.6        & 53.0        & 55.3        \\
\name (CSLS)               & 44.9        & 54.4        & 56.5       & 41.3        & 55.8        & 58.0       \\
\hline
\end{tabular}
\endgroup
}
\vspace{-3mm}
\end{table}

\section{Discussion and future work}
\label{sec:discussion}

\xhdr{Performance}
Several prior methods for \cl document embedding, including the previous state of the art \cite{conneau2017word}, work by first obtaining a monolingual word-level embedding for each language separately, then aligning the individual monolingual embedding spaces to each other in a postprocessing step, and finally heuristically averaging word vectors to obtain document vectors.
This approach relies on the quality of monolingual embeddings, which is often poor for low\hyp resource languages.
Our method, \name, on the contrary, is trained to directly obtain high\hyp quality document vectors.
It is more data\hyp efficient by leveraging document\hyp level alignment as a weak supervised signal, which results in superior \cl document representations, as evident in the large improvements we achieve on a \cl document retrieval task (\Secref{sec:doc retrieval}).

Moreover, we show that, although \name is trained on Wikipedia documents, it also generalizes to much shorter texts (sentences) whose content is of a very different nature (parliament debates) (\Secref{sec:sentence retrieval}).
The columns of our embedding map $\mPhi$ (\Eqnref{eq:W_definition}) may be interpreted as word embeddings,
and we show that, although not quite as effective as state-of-the-art word embeddings trained specifically as such, we nonetheless also achieve competitive performance in a word retrieval task, outperforming most prior methods (\Secref{sec:word retrieval}).

\xhdr{Computational complexity}
In addition to performance, our method also has the advantage of being massively scalable by
relying exclusively on matrix--vector multiplications.
A na\"ive implementation of the steps outlined in \Secref{sec:Cr5} takes 15 hours for 4 languages (English, Italian, Danish, Vietnamese),
and 4 days for 28 languages,
but it can be massively sped up by noting that
matrix--vector multiplication with $\hat{\mX}^\top\hat{\mX} + \lambda \mI$
dominates the computation time of our procedure, since it is repeatedly used to solve equations in the eigenvalue decomposition of \Eqnref{eq:eig_matrix} and then in the computation of $\mPhi$ in  \Eqnref{eq:answer}.
Expediting these multiplications can help tremendously. In the case of crosslingual embedding, this matrix is nearly block\hyp diagonal in the sense that it may be expressed as the sum of a block\hyp diagonal matrix with a rank-1 matrix
$\hat{\mX}^\top\hat{\mX} + \lambda \mI = (\mX^\top \mX + \lambda \mI) - n\vmu \vmu^\top,$
where $\vmu := \frac{1}{n}\mX^\top\mathbf{1}$. In particular, the blocks of matrix $\mB = \mX^\top \mX + \lambda \mI$ correspond to different languages.
The Woodbury identity then allows us to express the inverse
$(\hat{\mX}^\top\hat{\mX} + \lambda \mI)^{-1} = \mB^{-1} - \vzeta \vzeta^\top$
as a block\hyp diagonal minus a rank-1 matrix, where $\vzeta := \sqrt{\frac{n}{n \vmu^\top \mB^{-1}\vmu - 1}}\mB^{-1}\vmu$. Once we have computed $\zeta$, computing $(\hat{\mX}^\top\hat{\mX} + \lambda \mI)^{-1}\uu$ amounts to computing $\mB^{-1}\uu$, which defines an \emph{independent} set of equations for each language and therefore trivially parallelizes all equation solving across languages.

\xhdr{Further loss functions}
While our general framework (\Secref{sec:rank-reduced classification}) can accommodate any loss function (\Eqnref{eq:multiclass}), this paper focuses on the squared loss as used in ridge regression.
Future work should explore further loss functions.
In particular, \Eqnref{eq:rlsc1} models both the few nonzeros, as well as the many zeros, of the extremely sparse class\hyp membership matrix $\mY$, although modeling the nonzeros is likely to be more important.
One way forward would be to use a ranking or a margin\hyp based loss instead of the squared loss.
Such more complex loss functions are generally not as efficiently optimized as our squared loss, however, so future work should investigate online optimization via stochastic gradient descent as a way forward.

\xhdr{Further future work}
We foresee numerous additional avenues for future research.
For instance, it would be interesting to look into richer feature representations than plain bags-of-words, \eg, based on $n$-grams.
Also, given that our embedding is trained on Wikipedia, future work should strive to apply it to Wikipedia\hyp specific applications, such as \cl passage alignment \cite{gottschalk2017multiwiki}, section alignment \cite{piccardi2018structuring}, and plagiarism detection \cite{potthast2011cross}.
Finally, while this paper uses only Wikipedia as a training corpus, training on different texts (\eg, sentences from the Europarl corpus) would be straightforward, and future work should aim to understand whether this can lead to better embeddings for given settings.


\section{Conclusion}
This paper provides a new view on \cl document embedding, by casting the problem as one of multiclass classification and deriving an embedding map by decomposing the learned weight matrix.
Experiments show that our method, \name, achieves state-of-the-art performance on a \cl document retrieval task, as well as competitive performance on sentence and word retrieval tasks.
We hope that future work will build on our framework to derive even more effective embedding schemes.


\bibliographystyle{ACM-Reference-Format}
\bibliography{bibliography.bib}

\end{document}